\documentclass[10pt,twocolumn,letterpaper]{article}

\usepackage[pagenumbers]{cvpr} % To force page numbers, e.g. for an arXiv version

\usepackage[dvipsnames]{xcolor}

\definecolor{cvprblue}{rgb}{0.21,0.49,0.74}
\usepackage[pagebackref,breaklinks,colorlinks,citecolor=cvprblue]{hyperref}
\usepackage{xcolor} % For coloring text
\usepackage{textcomp}

\usepackage{makecell}
\usepackage[most]{tcolorbox}
\makeatletter
\newcommand{\ccell}[3][]{%
  \kern-\fboxsep
  \if\relax\detokenize{#1}\relax
    \expandafter\@firstoftwo
  \else
    \expandafter\@secondoftwo
  \fi
  {\colorbox{#2}}%
  {\colorbox[#1]{#2}}%
  {#3}\kern-\fboxsep
}
\makeatother
\definecolor{valhl}{HTML}{d9ead3}
\newcommand{\hl}[1]{\ccell{valhl}{#1}}

\def\paperID{*****} % *** Enter the Paper ID here
\def\confName{CVPR}
\def\confYear{2024}

\title{LLaVA-Gemma: Accelerating Multimodal Foundation Models with a Compact Language Model}

\author{Musashi Hinck\footnotemark[1] \and Matthew L. Olson\footnotemark[1] \and David Cobbley \and Shao-Yen Tseng \and Vasudev Lal\\
Cognitive AI, Intel Labs\\
Santa Clara, CA USA\\
{\tt\small {\{musashi.hinck,matthew.lyle.olson,david.j.cobbley,shao-yen.tseng,vasudev.lal\}}@intel.com}
}

\begin{document}
\maketitle
\renewcommand*{\thefootnote}{\fnsymbol{footnote}}
\footnotetext[1]{Equal Contributions, order decided by LLaVA-Gemma 2b}
\renewcommand*{\thefootnote}{\arabic{footnote}}

\begin{abstract}
% We train a suite of multimodal foundation models using the recently released Gemma models and the LLaVA framework. 
% In the burgeoning field of multimodal foundation models (MMFMs), the trade-off between computational efficiency and model performance remains a pivotal challenge. This paper introduces LLaVA-Gemma, a novel framework that leverages a smaller language model, Gemma, with 2 billion parameters, as opposed to the traditional use of larger models like LLaMA with 7+ billion parameters. Our primary contribution lies in the significant acceleration of multimodal tasks, achieving a XX\% reduction in inference time while only suffering a modest XX\% decrease in performance across various visual language tasks. We publicly release the code and weights for both 2b and 7b versions of our model.
% \matt{Main framing: Gemma is an important model as its the only one released so small with a 7b comparison}
% To facilitate more open research into multimodal foundation models, we will release trained checkpoints, as well as training and evaluation scripts

We train a suite of multimodal foundation models (MMFM) using the popular LLaVA framework with the recently released Gemma family of large language models (LLMs). Of particular interest is the 2B parameter Gemma model, which provides opportunities to construct capable small-scale MMFMs. In line with findings from other papers in this space, we test the effect of ablating three design features: pretraining the connector, utilizing a more powerful image backbone, and increasing the size of the language backbone. The resulting models, which we call LLaVA-Gemma, exhibit moderate performance on an array of evaluations, but fail to improve past the current comparably-sized SOTA models. Closer analysis of performance shows mixed effects; skipping pretraining tends to reduce performance, larger vision models sometimes improve performance, and increasing language model size has inconsistent effects.  We publicly release training recipes, code and weights for our models for the LLaVA-Gemma models\footnote{\url{https://huggingface.co/intel/llava-gemma-2b/}, \url{https://huggingface.co/intel/llava-gemma-7b/} }.

\end{abstract}

\vspace{-2ex}
\section{Introduction}
\vspace{-1ex}
In this paper, we introduce LLaVA-Gemma, a suite of vision-language assistants trained from the Gemma Large Language Model (LLM) variants, Gemma-2B and Gemma-7B \cite{team2024gemma}. Our work is inspired by the rapid progress in small but capable visual language models (VLMs), such as LLaVA-Phi \cite{zhu2024llava}, which have demonstrated remarkable efficiency and effectiveness in various language understanding tasks. LLaVA-Gemma distinguishes itself among small VLMs due to the public release of similarly trained, different-sized LLMs Gemma-2B and Gemma-7B. 

The unique release of the Gemma models offers an opportunity to contrast model performance in relation to parameter size and visual encoding capabilities. By possessing two variants with different parameter sizes, LLaVA-Gemma allows researchers to investigate the trade-offs between computational efficiency and the richness of visual and linguistic understanding. With these two variants, we perform a deeper exploration of how varying levels of model complexity influence the effectiveness of visual encoding, providing valuable insights into the optimization of small VLMs for diverse tasks and environments. Furthermore, the use of significantly more unique tokens, at $256k$, offers an opportunity to investigate how a massively increased token set effects multi-modal performance.

% Our work builds into three streams. The increased availability of powerful LMMs such as GPT-4V and Gemini \mh{cite} has created novel applications and research streams bridging modalities through a unified language and image interface. Work on 
%On the larger end of the spectrum, commercial LMMs such as GPT-4V \cite{achiam2023gpt} and Gemini \mh{cite} provide users a means of bridging modalities through a unified language and image interface. Recent work has found effective and efficient frameworks for combining existing open-source models to produce capable LMMs \citep{li2023multimodal,liu2023improved}.

Recent advancements in (LLMs) \cite{vaswani2017attention} and multimodal foundation models (MMFMs) \cite{li2023multimodal} have propelled the interest and development of Large Multimodal Models (LMMs).
Notable models like GPT-4 \cite{achiam2023gpt}, LLaVA \cite{liu2024visual,liu2023improved}, and their derivatives have demonstrated significant performance in vision-language tasks such as Visual Question Answering (VQA) and image captioning \cite{hudson2019gqa}. However, the computational demands of deploying these models have led to the exploration of small-scale LMMs.
Our work aims to provide a unified analysis of small-scale LMMs, examining how model selections, training recipes, and data contribute to performance, which is distinct from existing works such as LLaVA-Phi.

Our contributions are as follows: 
\begin{enumerate}
\item We introduce LLaVA-Gemma, a MMFM that leverages the compact yet powerful Gemma language models for efficient multimodal interactions.
\item We extensively evaluate Gemma-2B and Gemma-7B model variants provides valuable insights into the trade-offs between computational efficiency and the richness of visual and linguistic understanding in LLMs.
\item We present a deep exploration into alternate design choices and visualize attention with relevancy maps to enhance our understanding of the model's performance and attention.

\end{enumerate}

\setlength\tabcolsep{3pt}
\begin{table*}[t]
\centering
%\resizebox{\textwidth}{!}
{
\begin{tabular}{llc | ccccccccc}
\toprule
\textbf{Language} & \textbf{Vision} & \textbf{Pretrain}  & & \multicolumn{2}{c}{\textbf{MME}} & \textbf{MM-} & \multicolumn{2}{c}{\textbf{POPE}} & & &  \textbf{ScienceQA} \\
\textbf{Backbone} & \textbf{Backbone} & \textbf{Connector} & \textbf{GQA} & \textbf{Cog.} & \textbf{Per.} & \textbf{Vet} & \textbf{Acc.} & \textbf{F1} & \textbf{VQAv2} & \textbf{MMVP} & \textbf{Image} \\
\midrule
\texttt{gemma-2b-it} & \texttt{CLIP}   & Yes & 0.531      & 236 & 1130 & 17.7 & 0.850 & \hl{0.839} & 70.7 & 0.287 & 0.564 \\
\texttt{gemma-2b-it} & \texttt{CLIP}  & No  & 0.481      & 249 & 935  & 13.1 & 0.784 & 0.762 & 61.7         & 0.180 & 0.549 \\
\texttt{gemma-2b-it} & \texttt{DinoV2} & Yes & \hl{0.587} & 307 & \hl{1133} & \hl{19.1} & \hl{0.853} & 0.838 & \hl{71.4} & 0.227     & 0.555 \\
\texttt{gemma-2b-it} & \texttt{DinoV2} & No  & 0.501      & \hl{309} & 959  & 14.5 & 0.793 & 0.772          & 61.7 & 0.180     & 0.568 \\
\texttt{gemma-7b-it} & \texttt{CLIP}   & Yes & 0.472      & 254 & 895  & 18.2 & 0.848 & 0.829 & 68.7        & \hl{0.327}     & 0.625 \\
\texttt{gemma-7b-it} & \texttt{CLIP}  & No  & 0.472      & 278 & 857  & 19.1 & 0.782 & 0.734 & 65.1         & 0.240   & \hl{0.636} \\
\texttt{gemma-7b-it} & \texttt{DinoV2} & Yes & 0.519 & 257 & 1021 & 14.3 & 0.794 & 0.762 & 65.2 & \hl{0.327} & 0.628 \\
\texttt{gemma-7b-it} & \texttt{DinoV2} & No  & 0.459 & 226 & 771 & 12.2 & 0.693 & 0.567 & 57.4 & 0.267 & 0.598 \\
\bottomrule
\texttt{Phi-2b}        & \texttt{CLIP}   & Yes & -          & -   & 1335 & 28.9 & -     & 0.850 & 71.4 & - & 0.684 \\
%\texttt{TinyLlava}  & \texttt{SigLIP}   & Yes & 0.620      & -   & 1465 & 32.0 & -     & 0.864 & 79.9 & 59.1 & 0.691 \\
\texttt{Llama-2-7b}    & \texttt{CLIP}   & Yes & 0.620      & 348 & 1511 & 30.6 & 0.850 & 0.859 & 78.5 & 46.1 & 0.704 \\
\bottomrule
\end{tabular}
\vspace{-2ex}
\caption{Performance of LLaVA-Gemma models across seven benchmarks. \hl{Highlighted box} indicates strongest performance amongst LLaVA-Gemma models. Bottom two rows show self-reported performance of Llava Phi-2 and LLaVA-v1.5 respectively.}
\label{tab:results}
}
\vspace{-3ex}
\end{table*}

%\section{Related Work}
%\input{documentBody/2_relatedwork}

\section{Methods}
\vspace{-2ex}
We follow the LLaVA framework \cite{liu2023improved} with a few design modifications. This framework combines a pretrained vision encoder (such as CLIP \cite{radford2021learning}) and pretrained language model (such as Llama-2 \cite{touvron2023llama}) into a multimodal model using a MLP connector and a two-stage training procedure.

The first stage pretrains the MLP connector by freezing the vision and language models and training on custom dataset of 595k samples filtered from CC3M \cite{sharma-etal-2018-conceptual}.
The second stage jointly finetunes the language model and connector using a custom mixture 665k multimodal instruction tuning examples. This dataset includes synthetic data generated \cite{liu2024visual}, as well as examples from established vision-language training sets such as GQA \cite{hudson2019gqa} and TextCaps \cite{sidorov2020textcaps}.

We deviate from the original recipe in three ways: the language model, the vision encoder and the pretraining stage.
For the language backbone, we use the recently released Gemma models \cite{team2024gemma}. Two aspects of Gemma make it an interesting candidate for our experiments. Whereas LLaVA uses the 7 and 13-billion parameter vicuña langauge models \cite{zheng2023judging}, Gemma offers 2 and 7-billion parameter versions. Next, Gemma uses a significantly larger token set than any other LLM, with 256k unique tokens (compared to a standard ~50k), which offers a unique opportunity to see the effects of a massively more diverse embeddings space.
Other papers exploring the design space of Vision Language Models (VLMs) find the vision encoder is important for achieving strong performance \cite{mckinzie2024mm1}. Correspondingly, we explore the use of the larger 1-billion parameter DINOv2 image encoder \cite{oquab2023dinov2} as the vision tower.
Related work on VLMs \cite{karamcheti2024prismatic} finds that skipping the initial pretraining stage improves downstream performance. For all designs, we train a version with and without the initial pretraining step.

\vspace{-2ex}
\section{Results}
\vspace{-2ex}
We evaluate the LlaVA-Gemma models on a similar collection of benchmarks to other LMM works: \textbf{GQA} \cite{hudson2019gqa}; \textbf{MME} \cite{fu2023mme}; \textbf{MM-Vet} \cite{yu2023mm}; \textbf{POPE} (accuracy and F1) \cite{li2023evaluating}; \textbf{VQAv2} \cite{goyal2017making}; \textbf{MMVP} \cite{tong2024eyes}; the image subset of \textbf{ScienceQA} \cite{lu2022learn}. Our experiments provide insights into the efficacy of various design choices within the LLaVA framework. As shown in table \ref{tab:results}, the performance of LLaVA-Gemma models across seven benchmarks reveals interesting patterns, particularly concerning the choice of vision encoder and the impact of pretraining the connector. 

One item of note is that for the ScienceQA dataset, the larger models consistently perform better than smaller due to the datasets task requiring diverse general knowledge captured better by the larger models.

\subsection{Influence of Vision Encoder on Performance}
\vspace{-1ex}
%The choice of vision encoder shows a significant impact on performance. The gemma-2b-it model with the DinoV2 encoder outperforms its CLIP counterparts in most benchmarks, achieving the highest scores in MME-Per. (1133), VQAv2 (71.4), and a competitive POPE F1 score (0.853). Notably, the model also obtains the top GQA score (0.587) among LLaVA-Gemma configurations, underscoring the effectiveness of DinoV2 in multimodal contexts.
For the 2B backbone, exchanging the CLIP vision encoder for DinoV2 appears to generally improve performance, with DinoV2 variants outperforming CLIP variants on all benchmarks except POPE-F1 and MMVP. When using a 7B backbone, the picture is murkier; although we see improvements for GQA and MME, we see a decline in performance on MM-Vet, POPE, VQA and ScienceQA. This may suggest an interaction between the capability of the language model and the richness of the representation provided by the vision encoder, or to the possibility that the 7b-Dino combination is undertrained.

\subsection{Effects of Pretraining}
\vspace{-1ex}
%Pretraining appears to play a crucial role in model performance, with pretrained models consistently outperforming their non-pretrained versions. For example, in the Gemma-2b-it CLIP configuration, pretraining leads to better performance in MME benchmarks like Cognitive (236 vs. 249) and Perceptual (1130 vs. 935), as well as in POPE Accuracy (0.850 vs. 0.784) and F1 (0.839 vs. 0.762). These results do not support the hypothesis posited by \citet{karamcheti2024prismatic}, as our observations show pretraining contributes positively to the fine-tuning phase.
We find that skipping the initial connector pretraining almost always reduces model performance. With the exceptions of 2B-Dino on MME Cognition and 7B-CLIP on MME Cognition, MM-Vet and ScienceQA, the variant with a pretrained connector outperforms its counterpart that skipped pretraining. These results do not support the hypothesis posited in \citet{karamcheti2024prismatic}.

\subsection{Comparison to Baselines}
\vspace{-1ex}
Contrasting the results of LLaVA-Gemma with the self-reported performances of Phi-2b and Llama-2-7b models provides additional context. The LLaVA-Gemma models only reach parity on comparably-sized baselines for the VQA benchmark between 2B models. Given the absence of strong \textit{a priori} reasons to expect Gemma-based LLaVA models to perform worse, understanding this ``poor'' performance is a direction of future interest.
%The Phi-2b and Llama-2-7b models report higher MM-Vet scores than any LLaVA-Gemma configurations, which may be attributed to differences in language backbones or training strategies.
%The ScienceQA Image benchmark shows a mixed response to pretraining, with the non-pretrained Gemma-7b-it model achieving the strongest performance (0.636) among the LLaVA-Gemma models, yet falling short when compared to the performance of the Llama-2-7b model (0.704).

\vspace{-1ex}
\subsection{Speed of Training and Inference}
\vspace{-1.5ex}

We compare the training and eval speed for the two models sizes.  In our experiments, the training time for the Gemma-2B model on 8 Intel Gaudi 2\textsuperscript{\tiny{\circledR}} AI accelerators was 4 hours, while the larger Gemma-7B model required 16 hours to train under the same conditions. This indicates that the Gemma-7B model, with its increased parameter count, takes approximately four times longer to train compared to the Gemma-2B model. The relative speed of the Gemma-7B model is thus 0.25x compared to the Gemma-2B model. We find a similar speed ratio during inference. These results highlight the trade-off between model size and training efficiency, with larger models requiring significantly more computational resources. 

%MH: shall we mention here that the 2B model was trained on Intel (c) Gaudi (c) 2 Accelerators?

\vspace{-1ex}
\section{Analysis}
\vspace{-1.5ex}

\subsection{Impact of Alternate Design Choices}
\vspace{-1ex}
\begin{figure}[t]
\centering
\includegraphics[width=0.99\linewidth]{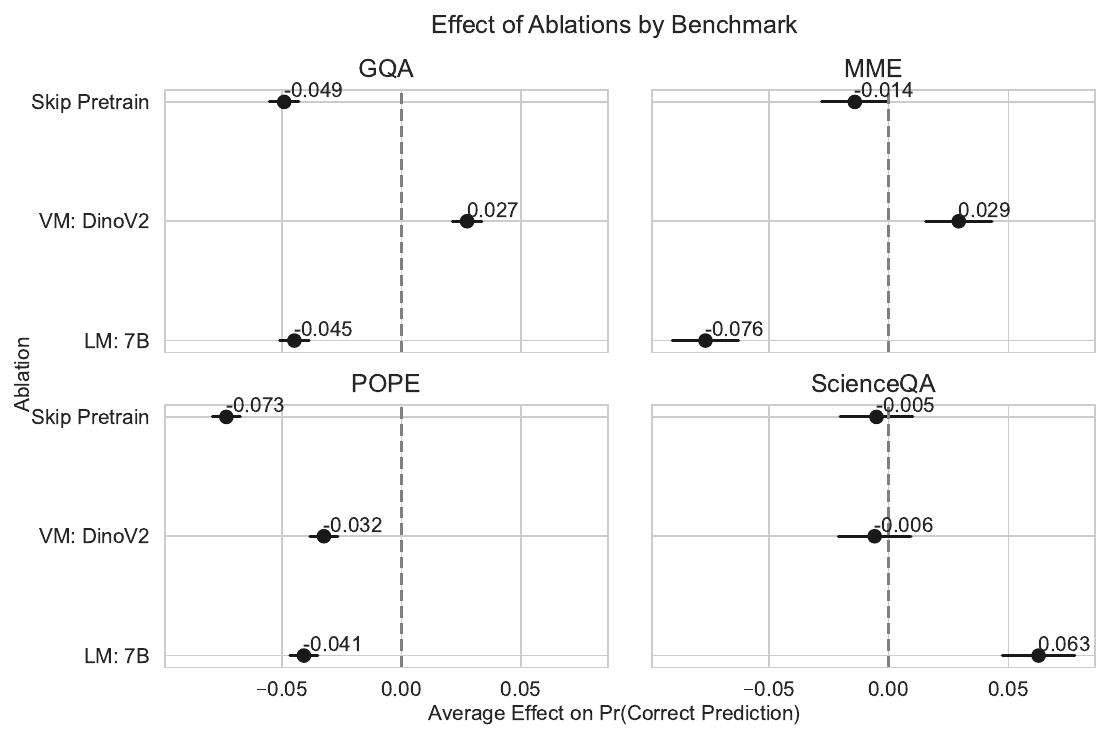}
\vspace{-2ex}
\caption{\textit{Effect of design choices differs between evaluations.} Point indicates average change in probability of correct answer versus baseline design.}
\label{fig:ablation}
\vspace{-3ex}
\end{figure}

Table \ref{tab:results} suggests that the \texttt{gemma-2b-dino} recipe generally provides stronger evaluation results, but these results are mixed. To better assess the effect of the design choices, we fit a collection of linear models to measure the average associated change in the probability of a correct prediction as a function of each of the three ablations: skipping pretraining, changing the vision backbone, and increasing the size of the LM backbone from 2B to 7B. We study these effects separately for each benchmark.

Figure \ref{fig:ablation} shows the average effects of design choices for four benchmarks where we have observation-level errors. Skipping pretraining appears to either have a strong negative (GQA, POPE) or weak/insignificant effect (MME, ScienceQA). Changing the vision encoder to DinoV2 improves performance on GQA and MME, but slightly worsens performance on POPE and has no significant effect on the probability of correct predictions on ScienceQA. Notably, in our experiments increasing the LM backbone to the 7B parameter variant had a strong negative effect on GQA, MME and POPE, but strong positive effect on ScienceQA. Taken together, these heterogeneous results underscore the need for more granular analysis of errors and design choices.
%We note that skipping pretraining either has no effect (MME, ScienceQA) or a significant negative effect (GQA, POPE); using DinoV2 improves performance on GQA but nothing else; and using the 7B Gemma backbone improves ScienceQA but decreases performance on GQA and MME. The heterogeneity in these results is tantalizing, and shows that a deeper investigation of the effect of these design decisions is required.

\vspace{-1ex}
\subsection{Visualizing Attention with Relevancy Maps}
\vspace{-1ex}
To better understand the differences between our the LLaVA-Gemma models, we use relevancy maps \cite{Chefer_2021_ICCV} to visualize where the model focuses its attention. 
These relevancy maps provide a token-wise understanding of the model's attention by highlighting the most relevant parts of the input and is specially designed to maintain the total relevancy across layers for transformer based models.

We apply an qualitative example of these relevancy maps from the Eyes-wide-shut (MMVP) dataset. 
This dataset is of particular interest as it is designed to find image-caption pairs that a CLIP model finds to be similar, but are distinct. 
As the traditional LLaVA recipe uses CLIP, we compare our CLIP backboned models to find a case where the Gemma 2b model fails, but Gemma 7b is successful.

\begin{figure}[bt]
\centering
\includegraphics[width=0.99\linewidth]{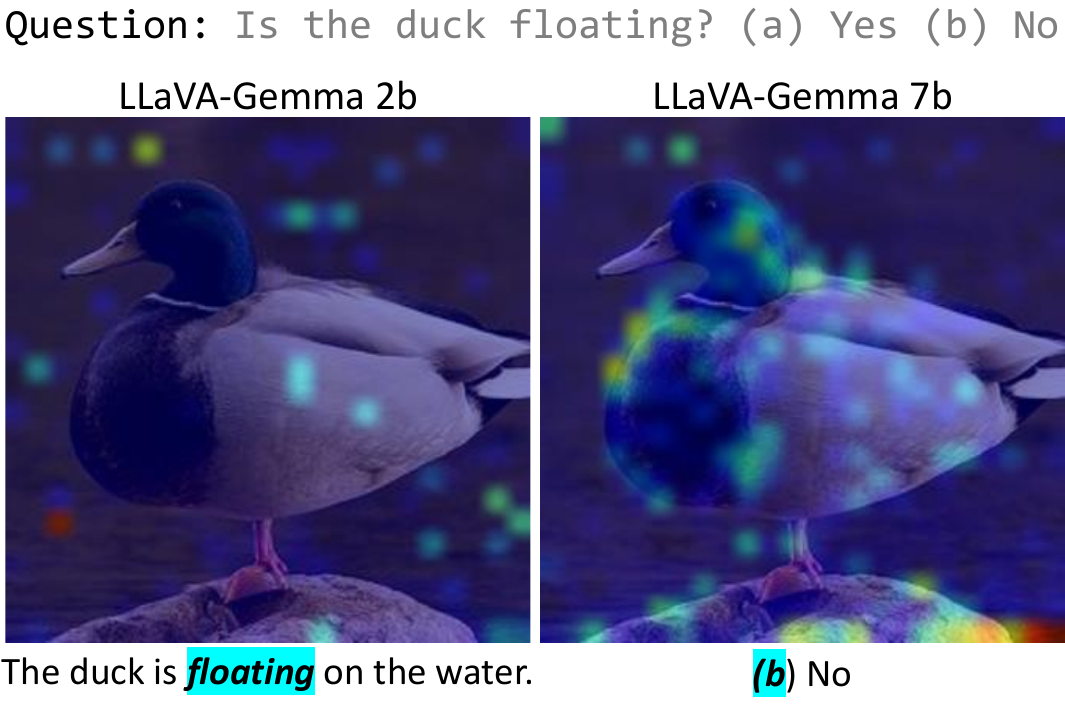}
\vspace{-2ex}
\caption{Relevancy map comparison between LLaVA-Gemma 2b (Left) and LLaVA-Gemma 7b (Right) with gradients on the first relevant output token. For the question ``Is the duck floating? (a) Yes (b) No'', despite using the identical CLIP vision encoder, the smaller model does not attend to the visual input.}
\label{fig:relevancy}
\vspace{-3ex}
\end{figure}

Figure \ref{fig:relevancy} shows an example of the differences in attention to the visual aspects of the scene between the LLaVA-Gemma 2b and LLaVA-Gemma 7b models. 
The relevancy maps for the LLaVA-Gemma 2b model show a dispersed and unfocused pattern of attention, which correlates with its failure to accurately interpret the scene. In contrast, the LLaVA-Gemma 7b model exhibits a more concentrated and relevant pattern of attention, particularly focusing border between objects: the duck, the water, and the rock being stood on.
This visualization not only highlights the superior performance of the LLaVA-Gemma 7b model, but also illuminates an interesting case where leveraging a more powerful LLM ensures improved visual token attention.

\section{Discussion}
\vspace{-1ex}
In this paper, we introduced LLaVA-Gemma, a compact vision-language model leveraging the Gemma Large Language Model in two variants, Gemma-2B and Gemma-7B. Our work provides a unique opportunity for researchers to explore the trade-offs between computational efficiency and multimodal understanding in small-scale models. The availability of both variants allows for a comparative analysis that sheds light on how model size impacts performance in various tasks. Our evaluations demonstrate the versatility and effectiveness of LLaVA-Gemma across a range of datasets, highlighting its potential as a benchmark for future research in small-scale vision-language models. With these models, future practitioners can optimize the performance of small-scale multimodal models more directly.
\vspace{-4ex}
{
    \small
    \bibliographystyle{ieeenat_fullname}
    \bibliography{00-main}
}


\end{document}